\documentclass[conference]{IEEEtran}
\IEEEoverridecommandlockouts
\usepackage{cite}
\usepackage{url}
\usepackage{amsmath,amssymb,amsfonts}
\usepackage{algpseudocode}
\usepackage{graphicx, subcaption}
\usepackage{textcomp}
\usepackage{xcolor}
\usepackage{wrapfig}
\usepackage{float}
\usepackage{booktabs}
\usepackage{caption}
\usepackage{times} 
\def\BibTeX{{\rm B\kern-.05em{\sc i\kern-.025em b}\kern-.08em
    T\kern-.1667em\lower.7ex\hbox{E}\kern-.125emX}}

\begin{document}
\title{Privacy-Preserving Optimal Parameter Selection for Collaborative Clustering}
%
%
\author{\IEEEauthorblockN{1\textsuperscript{st} Maryam Ghasemian}
\IEEEauthorblockA{\textit{Department of Computer and Data Sciences} \\
\textit{Case Western Reserve University}\\
Cleveland, OH, USA \\
Maryam.ghasemian@case.edu}
\and
\IEEEauthorblockN{2\textsuperscript{nd} Erman Ayday}
\IEEEauthorblockA{\textit{Department of Computer and Data Sciences} \\
\textit{Case Western Reserve University}\\
Cleveland, OH, USA \\
Erman.ayday@case.edu}
}
%
%
\maketitle
\begin{abstract}
This study investigates the optimal selection of parameters for collaborative clustering while ensuring data privacy. We focus on key clustering algorithms within a collaborative framework, where multiple data owners combine their data. A semi-trusted server assists in recommending the most suitable clustering algorithm and its parameters. Our findings indicate that the privacy parameter ($\epsilon$) minimally impacts the server’s recommendations, but an increase in $\epsilon$ raises the risk of membership inference attacks, where sensitive information might be inferred. To mitigate these risks, we implement differential privacy techniques, particularly the Randomized Response mechanism, to add noise and protect data privacy. Our approach demonstrates that high-quality clustering can be achieved while maintaining data confidentiality, as evidenced by metrics such as the Adjusted Rand Index and Silhouette Score. This study contributes to privacy-aware data sharing, optimal algorithm and parameter selection, and effective communication between data owners and the server.
\end{abstract}
%
%
%
\section{Introduction}\label{Introduction}
Clustering, a fundamental technique in unsupervised machine learning, involves identifying patterns in unlabeled data. This process includes feature selection, measuring data similarity, and evaluating algorithms~\cite{jain1999data,xu2005survey}. There are several types of clustering algorithms: partitioning based~\cite{wu2021secure,mohassel2020practical}, distribution based~\cite{hamidi2019privacy,lin2005privacy}, density based ~\cite{bozdemir2021privacy,anikin2017privacy,rahman2017towards}, and hierarchical~\cite{meng2019private,de2014secure}. Our study concentrates on selecting the optimal hyperparameters for key representative clustering algorithms from each category, within a privacy-preserving collaborative framework. Specifically, we explore K-Means (partitioning-based), Hierarchical Clustering (HC, hierarchical), Gaussian Mixture Models (GMM, distribution-based), and DBSCAN (density-based).Choosing the right parameters is crucial as it directly impacts the accuracy and effectiveness of the clustering results, thereby influencing the insights derived from the data while maintaining privacy.

Motivated by the fact that clustering algorithm perform better with larger amount of data and that datasets are typically distributed across different parties, cooperative clustering and collaborative clustering ~\cite{cornuejols2018collaborative} techniques have been popular. In cooperative clustering, each party generates its own clustering results, and a final clustering is performed via a post-processing step once individual processes are completed. In contrast, collaborative clustering aims to leverage the contributions of multiple parties by exchanging information about local data, current hypothesized clustering, or algorithm parameters to benefit each other's computations. Due to privacy concerns of the parties, privacy-preserving algorithms have been proposed during collaborative clustering, which aim to protect the sensitive information in each parties' local dataset. 
However, depending on the type of clustering (partitioning-based, distribution-based, density-based, or hierarchical clustering), parties need to decide on some common input parameters. Selection of such parameters significantly effect the accuracy of the clustering algorithm and existing privacy-preserving collaborative clustering techniques assume such parameters are pre-selected. On the other hand, such parameters typically depend on the distribution of the federated dataset of the parties and they should be determined in a privacy-preserving way before the collaborative clustering. In addition, parties also need to decide the type of the clustering algorithm depending on their federated dataset as different types of algorithms perform differently in particular datasets. 
To fill this gap, in this paper we focus on collaborative clustering under a server-assisted scenario. The goal is to evaluate the effectiveness of server-provided input parameters and the type of clustering algorithm in achieving optimal clustering results. We consider representative clustering algorithms from each of the aforementioned categories: K-Means as a partitioning-based algorithm, Hierarchical Clustering (HC) as a hierarchical algorithm, Gaussian Mixture Models (GMM) as a distribution-based approach, and Density-Based Spatial Clustering of Applications with Noise (DBSCAN) as a density-based algorithm. We conduct a series of experiments using a labeled dataset consisting of numeric data. The experiments involve calculating input parameters and assessing the clustering results using metrics such as the Adjusted Rand index (ARI) and Silhouette Score. 

The motivation for employing a semi-trusted third party, or cloud server, in this context stems from the dual benefits of enhancing privacy and facilitating the selection of optimal clustering algorithms and their parameters without necessitating large computational resources on the part of data owners. By leveraging differential privacy techniques, our approach aims to safeguard data privacy throughout the collaborative clustering process. The semi-trusted server plays a pivotal role in this framework by recommending the most suitable clustering algorithm and hyperparameters, thus streamlining the collaborative effort while prioritizing data confidentiality.
Our findings demonstrate the efficacy of this approach. We were able to determine optimal clustering algorithm and its hyper-parameter(s) that maintain data privacy and still deliver high-quality clustering, evidenced by metrics such as the Adjusted Rand Index and Silhouette Score. Also, it is found that the Randomized Response mechanism preserve the gap in data, showcasing its efficiency in upholding the data's original structure while implementing privacy protection measures.

In this work, we make the following contributions to the context of collaborative clustering with hyper parameter recommendation:

    1. Privacy-Preserving and Efficient Communication: We introduce a novel privacy-preserving step in the collaborative clustering process, where data owners share parts of their datasets with the server after applying the randomized response (RR) mechanism to add noise to their respective datasets. This step enhances privacy protection by concealing sensitive information while still allowing for meaningful analysis. Additionally, we establish a seamless communication framework between the data owners and the server, ensuring privacy-preserving data sharing. 
    Unlike previous works that primarily rely on pre-selected clustering parameters and then apply encryption techniques in distributed or collaborative clustering, our approach goes beyond by addressing the challenge of parameter selection by determining the optimal clustering algorithm along with the respective hyper-parameters and incorporating the randomized response (RR) mechanism to introduce noise and safeguard sensitive information during data sharing. 

    2. Optimal Algorithm Selection: The server plays a crucial role in identifying the optimal clustering algorithm and its corresponding hyper-parameters. By employing various methods, the server evaluates different algorithms and provides data owners with recommendations for achieving the best clustering results. This step helps alleviate the burden of algorithm selection and parameter tuning for data owners.

    3. Server-Data Owner Interaction: The server communicates chosen algorithms and parameters back to the data owners, ensuring that all parties are aligned with the recommended strategies. This facilitates a coordinated effort that enhances both accuracy and efficiency.

In summary, our study contributes to privacy-aware data sharing, optimal algorithm and hyper-parameter selection, and effective communication between data owners and the server. The results revealed that the amount of noisy data shared and the privacy budget ($\epsilon$) did not significantly affect the server's algorithm and parameter recommendations. However, an increase in the privacy budget was found to elevate the risk of membership inference attacks, suggesting a trade-off between privacy protection and attack vulnerability.

\section{Related Work}\label{Related-Work}
In our study, we delve into clustering as an unsupervised machine learning technique for uncovering patterns in unlabeled data. We provide a concise review of privacy-preserving approaches in distributed and collaborative clustering, differentiated by their algorithm types as introduced in Section~\ref{Introduction}. Notably, existing methodologies predominantly utilize predefined algorithms and hyper-parameters, setting a stage for our novel contribution: dynamically identifying optimal clustering algorithms and hyper-parameters to enhance collaborative clustering performance in a privacy-aware manner.

Bi et al.'s PriKPM scheme ~\cite{bi2023outsourced} introduces a pioneering privacy-preserving k-prototype clustering method that utilizes additive secret sharing for handling mixed data types in cloud computing environments, addressing the critical privacy concerns inherent to cloud services. This framework splits data samples for secure processing by dual collaborative servers, ensuring clustering privacy through secure distance computation and cluster updates. Their secure initialization and comparison protocols underscore the model's effectiveness and security, validated through experiments demonstrating computational efficiency and maintained clustering accuracy.

Wang et al.'s work ~\cite{wang2023privacy} proposes a privacy-preserving k-means clustering model for the IoT ecosystem, leveraging multi-key fully homomorphic encryption to enable secure computations between edge nodes and the cloud. This model delineates a novel cloud-edge collaborative framework that optimizes resource use while preserving data privacy. It includes secure communication protocols essential for the k-means algorithm, showcasing the model's ability to protect data and model privacy with reduced overhead, proving the viability of privacy-sensitive cloud-edge collaborations in IoT settings.

Further contributions include Jagannathan and Wright's ~\cite{jagannathan2005privacy}, as well as Baby et al.'s  ~\cite{baby2016distributed}, protocols for privacy-preserving distributed K-Means clustering, designed for data partitioned arbitrarily. These protocols maintain data confidentiality while following the K-Means algorithm's iterative nature, allowing secure computation of cluster centers and distances without data exposure.

Additionally, Lin et al. ~\cite{lin2005privacy} present an expectation maximization-based strategy for private clustering across distributed sites, utilizing secure summation to protect horizontally partitioned data. Liu et al. ~\cite{liu2013privacy} offer privacy-preserving DBSCAN techniques for data distributed in various ways, employing a Multiplication protocol based on additive homomorphic encryption for secure clustering.

Meng et al. ~\cite{meng2019private} introduce privacy-preserving hierarchical clustering algorithms, emphasizing a two-party model that employs homomorphic encryption and garbled circuits. Their approach provides a dendrogram depicting the clustering process, enriched with detailed merge metadata.

These diverse approaches share a common goal of enhancing privacy in collaborative clustering, yet they employ fixed algorithms and parameters. Our study seeks to advance this domain by focusing on adaptive parameter selection to achieve optimal clustering results, reflecting a significant leap toward balancing privacy preservation and analytical utility in collaborative settings.

\section{Background}
In this section we review some background and definitions of different clustering algorithms and clustering evaluation metrics as well as the local differential privacy. 
\subsection{Clustering Algorithms}\label{clustering_algorithms}
This study examines four clustering algorithm types: partitioning based, distribution based, density based, and hierarchical~\cite{wu2021secure,mohassel2020practical,bozdemir2021privacy,meng2019private,hamidi2019privacy,lin2005privacy,anikin2017privacy,de2014secure}. K-Means  algorithm is one of the simplest and most widely used unsupervised learning algorithms for clustering. It partitions the dataset into K distinct, non-overlapping clusters by minimizing the sum of distances between the data points and their respective cluster centroid, which serves as the representative of the cluster~\cite{wu2021secure,mohassel2020practical}. Gaussian Mixture Models (GMM) accommodates clusters with different sizes and correlations by assuming that the data points are generated from a mixture of several Gaussian distributions. Each cluster corresponds to a different Gaussian distribution, making this algorithm highly flexible when dealing with complex cluster structures~\cite{hamidi2019privacy,lin2005privacy}. DBSCAN stands out for its ability to identify clusters of arbitrary shapes and sizes, based on the density of the data points. It groups together closely packed points and marks points in low-density regions as outliers. DBSCAN does not require the specification of the number of clusters in advance. However, its performance is sensitive to the setting of its two parameters: the neighborhood size (eps) and the minimum number of points required to form a dense region (minPts)~\cite{bozdemir2021privacy,anikin2017privacy}. Hierarchical clustering builds a tree of clusters and does not require a pre-specified number of clusters. Hierarchical clustering can be either agglomerative (bottom-up approach) or divisive (top-down approach). It is particularly useful for hierarchical data and can provide insights into the data structure at different levels of granularity. However, it is computationally expensive for large datasets and is not scalable. The results can also vary significantly based on the choice of the linkage criterion, which determines how the distance between clusters is measured~\cite{meng2019private,xu1998distribution,everitt2011cluster,ester1996density,fukunaga1975estimation}. 

\begin{table*}[h]
\centering
\caption{Table of symbols and notations. }
\label{tab:symbols_notations}
\begin{tabular}{|l|l|}
 \hline
Symbol & Description  \\
\hline
$D_i$ & Dataset of each data owner $i$\\
$ND_i$ & Noisy data of each data owner $i$ produced as a result of RR\\
$f_{NDi}$ & Portion of the noisy data, $ND_i$, shared with server from each data owner $i$  \\
RR & Randomized Response mechanism\\
$\epsilon$ , eps& epsilon, Privacy Parameter \\
Eps & Epsilon, Maximum distance between clusters in DBSCAN\\
k & Number of clusters \\
ARI & Adjusted Rand Index \\
CH & Calinski-Harabasz Index \\
Homo & Homogeneity of the clusters \\
Comp & Completeness \\
\hline
\end{tabular}    
\end{table*}
\subsection{Evaluation Metrics for Clustering Algorithms}
This section outlines the evaluation metrics used to assess the effectiveness of the proposed privacy-preserving collaborative clustering approach. To measure the performance of our approach, we use the following metrics, each selected for its capability to capture various dimensions of clustering quality and privacy preservation:

\textbf{Adjusted Rand Index (ARI):} The Adjusted Rand Index is a measure of the similarity between two data clusterings. An ARI score ranges from -1 to 1, where 1 signifies perfect agreement between the clusterings, 0 indicates random labeling, and negative values represent independent clusterings. \textit{Higher} ARI values are preferable, as they denote greater concordance between the clustering output and the true labels, thereby reflecting more accurate clustering performance.

\textbf{Silhouette Score:} This metric evaluates how well each object lies within its cluster, with a focus on cohesion and separation. The score ranges from -1 to 1, where a high value indicates that the object is well matched to its own cluster and poorly matched to neighboring clusters. \textit{Higher} Silhouette Scores signify better-defined clusters, making it a valuable metric for assessing the distinctiveness of the clusters formed by the algorithm.

\textbf{Calinski-Harabasz Index (CH):} The Calinski-Harabasz Index is a method for evaluating the quality of a clustering. It measures the ratio of the sum of between-clusters dispersion and of within-cluster dispersion for all clusters. The \textit{higher} the value of the CH index, the better the clustering model, as it indicates a higher degree of separation between clusters.


\textbf{Classification Accuracy:} We also added classification accuracy to our evaluation framework, a metric that measures the proportion of correct predictions. Although unusual in unsupervised learning tasks like clustering, it helps evaluate how well cluster assignments match predefined labels when known. This metric is key in scenarios with known data classifications, allowing for direct comparison between our privacy-preserving clusters and actual categories.


Table~\ref{tab:symbols_notations} contains a list of symbols and notations used throughout this paper.


\subsection{Local Differential Privacy and Randomized Response Mechanism} \label{local differential privacy}
Local Differential Privacy (LDP)~\cite{costello2015geppetto,davidson2011on} is a more restricted form of traditional differential privacy~\cite{davidson2011provenance}. Unlike traditional differential privacy, LDP does not rely on a trusted third party and provides a higher level of data protection for users. In LDP, each user modifies their own data before sharing them with a data aggregator. The aggregator only sees the perturbed data, ensuring privacy. 
An algorithm \(A\) satisfies $\epsilon$-local differential privacy ($\epsilon$-LDP) if, for any input values \(v1\) and \(v2\): 
\begin{math}
  Pr[A(v1)=y]\leq e^\epsilon Pr[A(v2)=y]  
\end{math},
This condition holds true for all possible outputs of the algorithm \(A\).
The randomized response mechanism is commonly used to achieve $\epsilon-LDP$ ~\cite{dey2011propub}. In this mechanism, an individual reports the true value of a single bit of information with probability \(p\) and flips the true value with probability \(1-p\), following the
\begin{math}
    (ln{\frac{p}{1-p}})-LDP
\end{math}
property. Although initially defined for binary inputs (e.g., yes/no), the randomized response mechanism can be generalized.
To achieve $\epsilon$-LDP, the generalized randomized response mechanism ~\cite{gymrek2013identifying} shares the correct value with probability 
\begin{math}
    p = \frac{e^\epsilon}{(e^\epsilon+m-1)}
\end{math}
where \textit{m} is the number of possible states. Each incorrect value is shared with the probability.
\begin{math}
    q = \frac{1}{(e^\epsilon+m-1)}
\end{math}. A data aggregator collects the perturbed values from individuals and aims to calculate the frequency of values in the population while preserving privacy.
\section{System and Threat Models}
In this section, we provide an explanation of the system and threat model for privacy-preserving hyper-parameter identification for collaborative clustering.

\subsection{System Model}
In the proposed system model, the party who aims to collaborate in clustering with other data owners is referred to as the ``data owner'' (or researcher), while the server represents a third party that assists the data owners in identifying the optimal clustering algorithm and hyper-parameters. Our approach focuses on the preliminary stages before actual clustering occurs in a collaborative environment. Our objective is to identify the optimal algorithm and input parameters for collaborative clustering among multiple data owners who wish to maintain data privacy.

As discussed, different types of clustering algorithms perform differently depending on the type and distribution of the datasets, and hence it is crucial to identify the optimal clustering algorithm type beforehand. In addition, in clustering algorithms like K-Means, hierarchical clustering (HC), and Gaussian mixture model (GMM), the number of clusters (referred to as (\textit{k})) is a critical hyper-parameter. In the case of DBSCAN, the key input parameters include the minimum cluster size (\textit{minpoint}) and the maximum distance between clusters (\textit{Eps}).

Once these optimal conditions are determined, clustering can then be executed using one of the existing algorithms mentioned in Section \ref{clustering_algorithms}. In this context, data owners selectively share differentially private data with a semi-trusted server. This server plays a crucial intermediary role, analyzing the noisy data to recommend the most suitable clustering algorithm and corresponding hyper-parameters for the data received from data owners.
\subsection{Threat Model}
In this section, we outline the considered threats in our proposed scheme, which involve both the server and the data owners. 

\textbf{Server}: In this study, the server is considered semi-honest, indicating it might engage in malicious activities, such as extracting sensitive information from the datasets of the individual parties (data owners), but it honestly follows the protocol execution. The server's role is pivotal, yet poses a risk of privacy violations. Privacy attacks like membership inference ~\cite{power2023sok,pyrgelis2020measuring,nergiz2007hiding}, deanonymization ~\cite{power2023sok,narayanan2008robust,narayanan2009deanonymizing}, and attribute inference ~\cite{power2023sok,dwork2017exposed} are concerns. Membership inference attacks aim to determine whether a specific record is in the dataset. Deanonymization attacks link anonymized data to actual identities using external information. Attribute inference attacks deduce sensitive attributes from observed data. In our setting, the most relevant is membership inference, where the server tries to determine if a specific record is part of one of the data owners' datasets, leading to privacy breaches. Our proposed scheme prevents this by sharing only a small, differentially-private portion of the dataset (\begin{math}
    f_{NDi}
\end{math}), which makes deanonymization more complex and significantly reduces the threat of membership inference. 

\textbf{Data Owners} \label{data-owners}: In our system model, we assume that the parties involved in the collaborative clustering are honest but curious. This means that while they trust each other and do not engage in malicious behavior, they may still be interested in learning about each other's data. This assumption is based on the fact that other literature (such as those in Section~\ref{Related-Work}) has already addressed the challenges posed by malicious or semi-honest data owners in collaborative clustering using privacy-enhancing techniques like homomorphic encryption. In our work, we specifically focus on the task of selecting the optimal algorithm and hyper-parameters for the clustering process. By concentrating on this aspect, we aim to improve the efficiency and effectiveness of collaborative clustering while assuming a cooperative environment among the data owners.

\section{Proposed Solution and Framework}
\begin{figure*}[h]
\centering
\includegraphics[width=\textwidth]{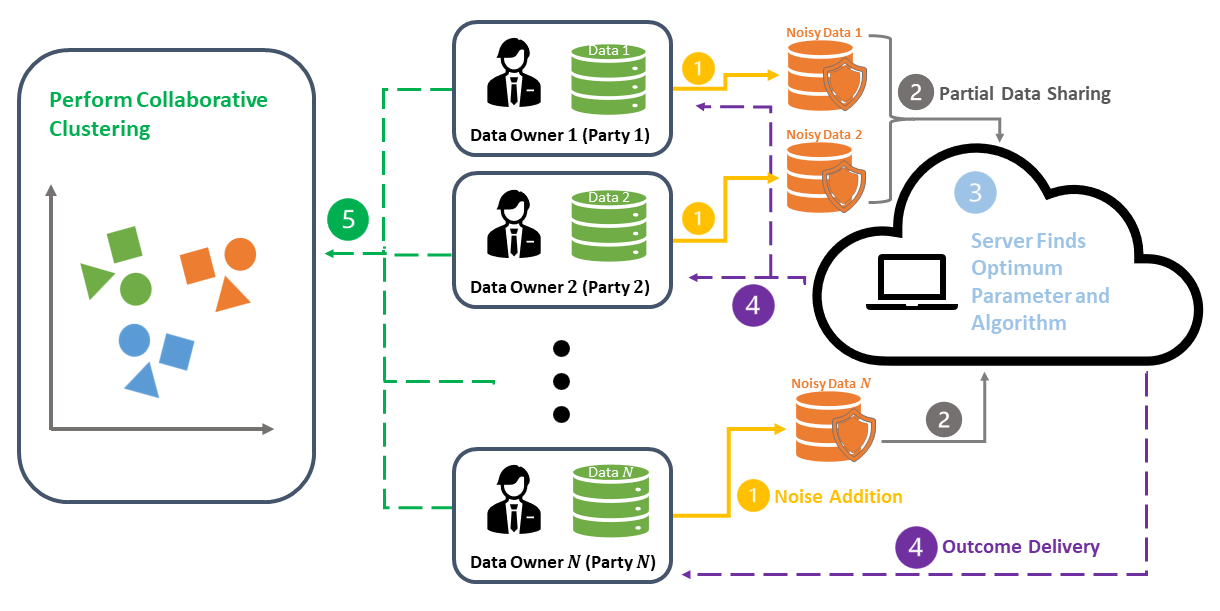}
\caption{Comprehensive five-step process, highlighting the interaction between multiple data owners and the server. We show how data are shared, processed for noise addition (to achieve differential privacy), and then utilized in a collaborative clustering algorithm, all while maintaining strict privacy protocols. In step (1), data owners add noise to part of their datasets using randomized response (RR). Data owners send a portion of their noisy data to the server in step (2). In step (3), the server applies various methods to find the optimum algorithm with its corresponding hyper parameter(s), and the server provides its outcome (algorithm and parameter) to the data owners in step (4). Finally, the data owners perform collaborative clustering based on server suggestions in step (5).}
\label{fig:SystemModel}
\end{figure*}

Our proposed system model and framework, as shown in Figure ~\ref{fig:SystemModel}, encompass five fundamental steps:

\textbf{Step 1-Noise Addition to Datasets}: Data owners (\begin{math}
    DO_{1\rightarrow N}
\end{math}) add noise to their datasets (to achieve differential privacy) through randomized response (RR) (\begin{math}
\{D_{1}, D_{2}, ..., D_{N}\} \rightarrow \{ND_{1}, ND_{2}, ..., ND_{N}\} 
\end{math}). 
In this process, we utilize a generalized version of the RR mechanism as mentioned in Section~\ref{local differential privacy}, allowing data owners to use perturbed data directly without encoding. The number of possible states for each feature (attribute) can vary according to the specific domain.

\textbf{Step 2-Data Sharing with the Server}: Data owners transmit a portion of their noisy data (\begin{math}
    f_{NDi}
\end{math}) to the server. During this step, data owners share their perturbed data with the server, enabling it to analyze the data and provide recommendations for the clustering process.

\textbf{Step 3-Server-Based Algorithm and Parameter Selection} \label{methods}: The server uses various methods to identify the best clustering algorithm and its hyper-parameters. In our previous discussions, we delved into the intricate process of collaborative clustering, a procedure involving two or more distinct parties, identified as data owners. A critical aspect of this arrangement is the confidentiality maintained between the parties; neither have insight into the data held by the other, thus upholding a strict privacy protocol. To further reinforce the privacy of their respective datasets, each data owner employs a technique known as the Generalized Randomized Response (RR) mechanism to achieve differential privacy during data sharing. This strategy is pivotal in introducing controlled distortions to their data, effectively masking the original information while preserving the overall structure and utility for analysis purposes.

Subsequent to the perturbation process, a subset of this noisy data from each data owner is transmitted to a semi-trusted server. The primary objective of this data sharing is to ascertain the most effective clustering algorithm and its associated input parameters, specifically tailored for the combined dataset. This step is crucial in optimizing the clustering process, ensuring that the combined data is analyzed in the most efficient manner possible. At this point, the server plays a key role. Upon receiving data from all parties, it merges these fragments into a singular, comprehensive dataset. Armed with this combined data, the server then embarks on a methodical exploration, employing techniques like the elbow method and the silhouette method. This exploration is aimed at pinpointing the most suitable input parameters for a variety of clustering algorithms, including K-Means, hierarchical clustering, Gaussian mixture models, and DBSCAN.

One of the challenges the server faced is the absence of ground truth data. To navigate this challenge, the server relies on performance evaluation metrics, such as the Silhouette Coefficient and the Calinski-Harabasz (CH) index. These metrics provide an objective measure of the efficacy of different algorithms, guiding the server in its selection process.

Here is the selection mechanism that server adapts to select the optimum clustering algorithm and its corresponding parameters for the data it received from data owners: 
The input to the selection algorithm includes a combined dataset from all data owners (\textit{data}), a list of candidate clustering algorithms (\textit{algorithms}), and a threshold parameter set to 0.1 ($\alpha$). The output is the optimal clustering algorithm (\textit{best\_algorithm}) and its corresponding parameters (\textit{best\_parameters}).

The procedure begins by initializing $\textit{max\_silhouette}$ to $-\infty$, 
$\textit{best\_algorithm}$ to \textit{None}, $\textit{best\_parameters}$ to \textit{None}, and $\textit{best\_ch\_index}$ to $-\infty$. The algorithm evaluates all candidate algorithms to find the maximum Silhouette score. For each algorithm and its parameter sets, the Silhouette score ($\textit{silhouette\_score}$) and Calinski-Harabasz index ($\textit{ch\_index}$) are calculated using the data. If $\textit{silhouette\_score} > \textit{max\_silhouette}$, $\textit{max\_silhouette}$ is updated to $\textit{silhouette\_score}$.

Next, the Silhouette score range is determined by setting $\textit{silhouette\_threshold}$ to $\textit{max\_silhouette} - \alpha$. The algorithm then selects the algorithms within this range that have the highest CH index. For each algorithm and parameter set, $\textit{silhouette\_score}$ and $\textit{ch\_index}$ are recalculated. If $\textit{silhouette\_threshold} \leq \textit{silhouette\_score} \leq \textit{max\_silhouette}$ and $\textit{ch\_index} > \textit{bes\_ch\_index}$, $\textit{best\_algorithm}$, $\textit{best\_parameters}$, and $\textit{best\_ch\_index}$ are updated. The final output is $\textit{best\_algorithm}$ and $\textit{best\_parameters}$.


For K-Means, HC, and GMM, the server focuses on determining the optimal number of clusters (\textit{k}) using the Elbow method and Silhouette Score ~\cite{yuan2019research}. The Elbow method looks at the point where the Within-Cluster-Sum of Squared Errors plateaus, while the Silhouette Score assesses cluster separation ~\cite{yuan2019research}. For DBSCAN, the (\textit{Eps}) value is set using the k-Nearest Neighbors algorithm, and the \textit{minpoint} parameter is adapted based on factors like data dimensionality, following different recommendations from prior research~\cite{ester1996density,sander1998density}.

\textbf{Step 4-Communication of Recommendations}: The server communicates the selected clustering algorithm type and its corresponding parameters to the data owners. At this stage, the server provides the data owners with the recommended algorithm and hyper-parameters based on the analysis of the shared data.

\textbf{Step 5-Execution of Collaborative Clustering}: Data owners use the suggested algorithm and hyper-parameters for collaborative clustering. As highlighted in Section ~\ref{Related-Work}, earlier methods in this area typically employed encryption for distributed or collaborative clustering. Our focus in this study, however, is not on executing clustering with privacy enhancement techniques; instead, we concentrate on the selection of the optimal algorithm and its corresponding hyper-parameters. Clustering is performed on the combined dataset under the assumption of mutual trust among the data owners. Further details on this are elaborated in Section~\ref{data-owners}. 

By following these steps, our framework provides recommendations for the optimal clustering algorithm and its hyper-parameters when data owners wish to perform clustering in a collaborative environment. 

\section{Evaluation}

\subsection{Datasets}


\textit{Dataset 1 (Obesity dataset)~\cite{uci2019obesity}}: This dataset, from the University of California Irvine Machine Learning Repository, consists of 2,111 records with 17 features, used to assess obesity levels based on diet and physical condition in Colombia, Peru, and Mexico. It classifies individuals into seven obesity levels and combines direct user data (23\%) with synthetic records (77\%) created using Weka and the SMOTE filter. The features include both numeric and categorical data.

\textit{Dataset 2 (Extended Iris dataset)~\cite{kaggle2023extended}}: It is an enhanced version of the classic Iris dataset~\cite{fisher1988iris}, featuring 1200 rows with 20 different attributes for each entry. It expands on the original by offering more detailed information, categorizing the rows into three classes for classification tasks. This dataset provides deeper biological and ecological insights about the iris flower, with its features divided into numeric and categorical types.

\subsection{Metric Significance and Evaluation Approach}

The combination of ARI, Silhouette Score, classification accuracy, and Calinski-Harabasz Index (CH) provides a comprehensive view of our approach's performance. ARI and Silhouette Score offer insights into the internal consistency and separation of clusters, respectively, while classification accuracy presents an external validation measure against known labels and the CH index complements these by assessing the ratio of between-cluster to within-cluster dispersion, highlighting cluster distinctness.
Together, these metrics enable a thorough assessment of both the clustering effectiveness and the impact of privacy-preserving techniques on data utility. In our evaluation, we analyze these metrics under varying conditions of data perturbation and privacy budget settings to explore the trade-offs between clustering quality and privacy preservation. The goal is to achieve optimal hyper-parameter selection that balances these aspects effectively, demonstrating the practical utility of our approach in collaborative clustering scenarios.

\subsection{Evaluation Results}
The datasets have undergone pre-processing to make them suitable for experiments. This includes converting categorical variables like 'Gender' in the dataset 1 (Obesity dataset) into numerical values. This encoding step transforms these variables into a format more conducive for analysis and computation in the later stages of the experiments, ensuring the datasets were optimally prepared for further exploration.

To determine the optimal number of clusters in our datasets, we use two methods: the elbow method and the silhouette method, as detailed in Section~\ref{methods}. Our experiments aim to find which method is most effective for our data's characteristics, particularly for different privacy budgets ($\epsilon$). The findings for the \textit{ dataset 1} using these methods are shown in Table~\ref{tab:Elbow_vs_Silhouette}.
Given that \textit{dataset 1} contains 7 clusters and \textit{dataset 2} contains 3, our analysis reveals that the elbow method is more effective than the silhouette method in determining the optimal cluster count. As a result, we employ the elbow method for a more in-depth analysis. This approach assists in identifying the optimal value of \textit{k} (number of clusters) for clustering algorithms such as K-Means, hierarchical clustering (HC), and Gaussian mixture models (GMM).

 \begin{table}[tb]
    \caption{Comparison of Silhouette and Elbow Methods for Predicting the Optimal Number of Clusters (\textit{k}): It highlights the superior performance of the Elbow method in predicting the optimal cluster count, leading to its selection for further analysis in this study.}
    \centering
    \begin{tabular}{|c|c|c|c|}
    \hline
         $\epsilon$ & Baseline K  & Silhouette K  &Elbow K \\
         \hline         
         0.0010 & 7 & 2 & 8\\
         0.1000 & 7 & 2 & 8\\
         1.00 & 7 & 2 & 8\\
         5.00 & 7 & 2 & 7\\
         10.00 & 7& 2 & 7\\
         \hline 
    \end{tabular}
    \label{tab:Elbow_vs_Silhouette}
\end{table}
\subsubsection{Optimum Input Parameter Selection Results on Noisy Datasets}
\begin{figure*}
\centering
\begin{subfigure}{0.45\textwidth}
    \includegraphics[width=\linewidth]{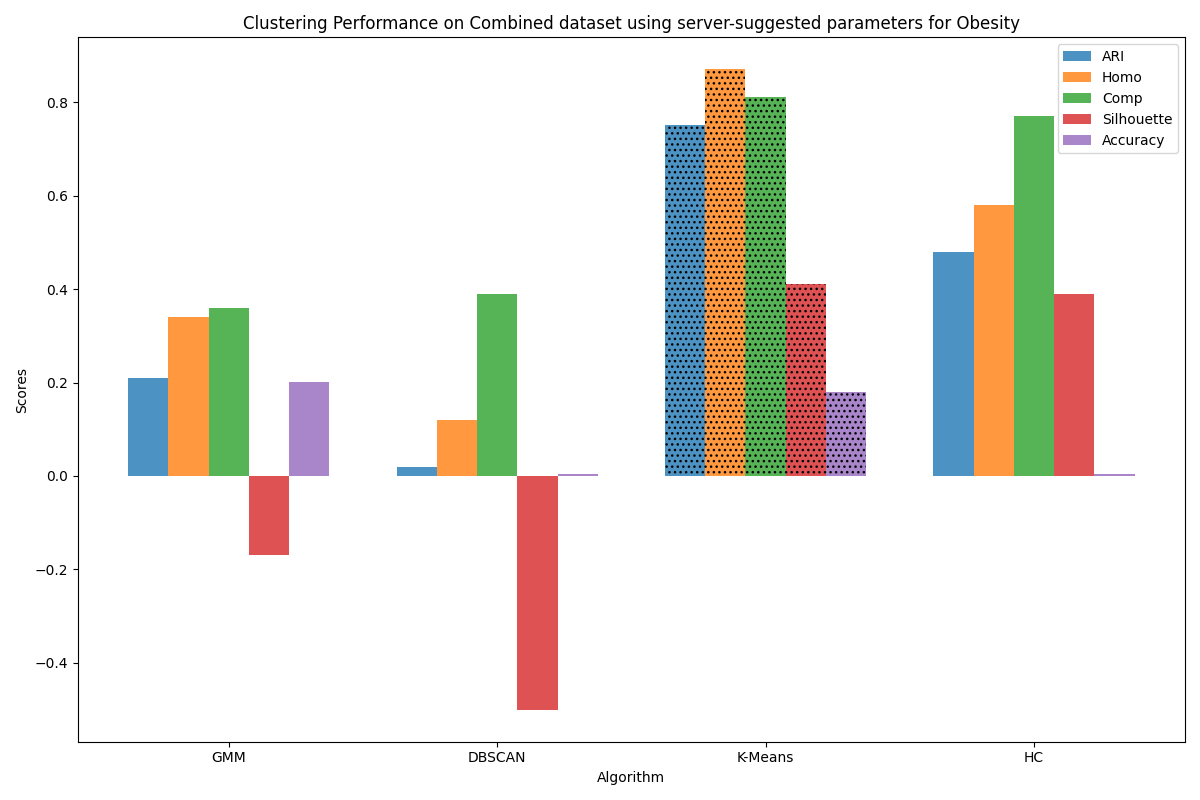}
    \caption{}
    \label{fig:clustering_Obesity}
    \end{subfigure}
\begin{subfigure}{0.45\textwidth}
    \includegraphics[width=\linewidth]{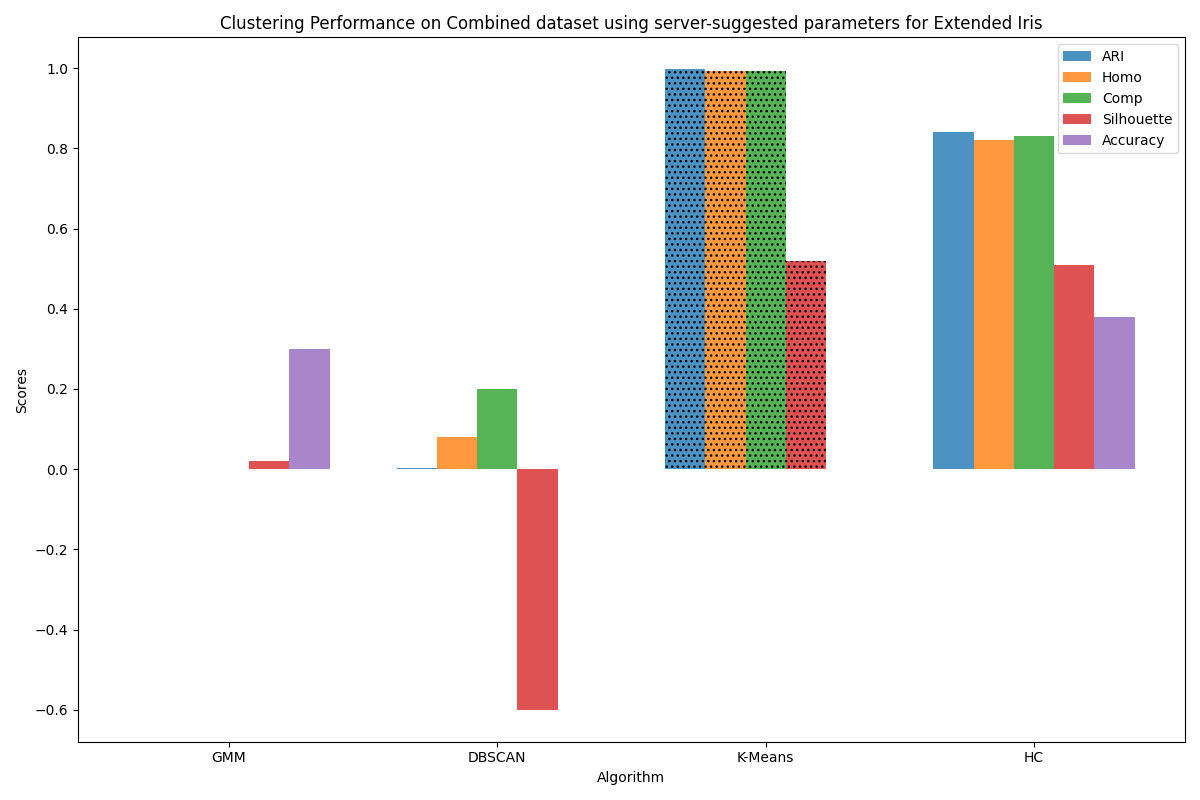} 
    \caption{}
    \label{fig:clustering_iris}
\end{subfigure}
    \caption{ Visual Representation of Clustering Algorithm Performance Across Combined Datasets. This figure illustrates the performance metrics from Table~\ref{tab:server_suggestion_10_0.1} for various clustering algorithms—GMM, DBSCAN, K-Means, and Hierarchical Clustering (HC)—evaluated under conditions of 10\% data sharing and a privacy parameter of $\epsilon = 0.1$. Performance metrics including Adjusted Rand Index (ARI), Homogeneity (Homo), Completeness (Comp), Silhouette Score, Calinski-Harabasz Index (CH), and Accuracy are plotted. Algorithms recommended by the server are highlighted with dots, showcasing their superior performance in comparison to others in each dataset scenario.}
    \label{fig: Clustering_on_Combined_dataset}
\end{figure*}

The experimental findings of this study are illustrated in Table~\ref{tab:server_suggestion_10_0.1} and Figure~\ref{fig: Clustering_on_Combined_dataset}. Table~\ref{tab:server_suggestion_10_0.1} offers a glimpse into the server's input parameter recommendations, based on the analysis of 10\% of the noisy data shared by the data owners, with a noise parameter ($\epsilon$) set at 0.1. Figure~\ref{fig: Clustering_on_Combined_dataset}, on the other hand, showcases the clustering outcomes derived from applying these server recommendations to the combined dataset. Notably, the results from this application highlight the superiority of the K-Means clustering algorithm for the combined dataset, a finding that resonates with the server's initial suggestion regarding the most suitable algorithm and hyper-parameter configuration.
These findings and recommendations by the server are not merely data points, but they serve as critical guidance for the data owners. They enable the owners to align their clustering strategies with the server's insights, which are rooted in a meticulous analysis of optimal input parameters. This alignment is key to enhancing the effectiveness and accuracy of the clustering process in a collaborative, privacy-preserving data environment.
\begin{table*}[ht]
    \caption{Server Suggestions for Clustering Input Parameters: Recommendations for various clustering algorithms based on 10\% shared noisy data ($\epsilon$ = 0.1). }
    \centering
    \begin{tabular}{|l|l|c|c|l|c|l|}
    \hline
         dataset & Algorithm & Data shared to Server & $\epsilon$ & K or Eps & Silhouette &CH \\
         \hline
          &GMM & 10\% & 0.1 & k = 8 & 0.34 & 301.30\\
          Dataset \#1 &DBSCAN &  10\% & 0.1 & k = 10, Eps = 1 & - & -\\
         &\textbf{K-Means} &  \textbf{10\%} & \textbf{0.1} & \textbf{k = 8} & \textbf{0.36} & \textbf{318.13}\\
         & HC &  10\% & 0.1 & k = 8 & 0.31 & 237.61\\
         \hline
          &GMM & 10\% & 0.1 & k = 3 &0.23 & 46.88 \\
         Dataset \#2 & DBSCAN & 10\% & 0.1 & k = 6, Eps = 7 & - & -\\
         & \textbf{K-Means} & \textbf{10\%} & \textbf{0.1} & \textbf{k = 3} & \textbf{0.36} & \textbf{61.92} \\
         & HC & 10\% & 0.1 & k = 3 & 0.37 & 51.57 \\
         \hline
    \end{tabular}
    \label{tab:server_suggestion_10_0.1}
\end{table*}


\subsubsection{Effect of Privacy Parameter $\epsilon$}
We have examined the influence of different levels of $\epsilon$, which perturb the data through the Randomized Response (RR) mechanism, on the server's ability to suggest input parameters for clustering algorithms. In this experiment, the server receives the same amount of data while varying the value of $\epsilon$, and its suggestions are evaluated on the joint dataset without any noise. 

\begin{table}[t]
    \caption{Differential Impact of Privacy Levels on Clustering Algorithms in the dataset \#1. This table explores the performance variations (measured through ARI, Silhouette, and Accuracy) of four distinct clustering algorithms (K-Means, HC, GMM, DBSCAN) at different privacy budget levels ($\epsilon$ = 0.1, 1, 5) with a consistent data sharing percentage (10\%).  }
    \centering
    \begin{tabular}{l|c|c|c|c|c|c}
    \hline
         Algorithm & Shared & $\epsilon$ & K  & ARI & Silhouette & Accuracy \\
         \hline
         K-Means & 10\% & 0.1 & k = 8 & 0.75 & 0.41 & 0.18 \\
         K-Means & 10\% & 1 & k = 8 & 0.75 & 0.41 &  0.18\\
         K-Means & 10\% & 5 & k = 7 & 1 & 0.44& 0.15 \\
         \hline
         HC & 10\% & 0.1 & k = 8 & 0.481 & 0.39& 0.005 \\
         HC & 10\% & 1 & k = 7 & 0.482 & 0.41 & 0.17\\
         HC & 10\% & 5 & k = 8 & 0.482 & 0.41 & 0.005\\
         \hline
         GMM & 10\% & 0.1 & k = 6 & 0.185 & -0.0143 & 0.201 \\
         GMM & 10\% & 1 & k = 8 & 0.2069 & -0.072 & 0.05\\
         GMM & 10\% & 5 & k = 6 & 0.2008 & -0.007 & 0.14 \\
         \hline
         DBSCAN & 10\% & 0.1 & k = 10 & 0.017 & -0.504 & 0.005\\
         DBSCAN & 10\% & 1 & k = 10 & 0.017 & -0.504 & 0.005\\
         DBSCAN & 10\% & 5 & k = 10& 0.017 & -0.504 & 0.005\\
    \end{tabular}
    \label{tab:effect_of_privacy_budget}
\end{table}
\begin{table}[b]
    \caption{Influence of Privacy Settings on Clustering Recommendations in the dataset \#2. This table details how varying privacy budgets ($\epsilon$ = 0.1, 1, 5) affect the recommendations for clustering parameters and subsequent algorithm performance (ARI, Silhouette, and Accuracy) for multiple clustering algorithms (K-Means, HC, GMM, DBSCAN), all with a consistent 10\% data sharing arrangement. }
    \centering
    \begin{tabular}{l|c|c|c|c|c|c}
    \hline
         Algorithm & Shared & $\epsilon$ & K & ARI & Silhouette & Accuracy \\
         \hline
         K-Means & 10\% & 0.1 & k = 3 & 0.997 & 0.52 & 0\\
         K-Means &10\% & 1 & k = 2 & 0.44 & 0.57& 0.18\\
         K-Means & 10\% & 5 & k = 3 & 0.997 & 0.52& 0 \\
         \hline
         HC & 10\% & 0.1 & k = 3 & 0.84 & 0.51& 0.38\\
         HC & 10\% & 1 & k = 3 & 0.84 & 0.51& 0.38\\
         HC & 10\% & 5 & k = 3 & 0.84 & 0.51& 0.38\\
         \hline
         GMM & 10\% & 0.1 & k = 3 & -0.0003 & 0.021 &0.3\\
         GMM & 10\% & 1 & k = 2 & -0.0004 & 0.051& 0.34\\
         GMM & 10\% & 5 & k = 3 & -0.0003 & 0.021& 0.3\\
         \hline
         DBSCAN & 10\% & 0.1 & k = 6 & 0.003 & -0.6 &0\\
         DBSCAN & 10\% & 1 & k = 6& 0.003 & -0.6 & 0\\
         DBSCAN & 10\% & 5 &  k = 6 & 0.003 & -0.6 &0 \\
    \end{tabular}
    \label{tab:effect_of_privacy_budget_iris}
\end{table}

Experimental results, as shown in Tables~\ref{tab:effect_of_privacy_budget} and~\ref{tab:effect_of_privacy_budget_iris}, reveal a notable consistency in the server's recommendations. Regardless of the $\epsilon$ value, the server consistently proposes around 7 clusters for the first dataset (Obesity dataset) and approximately 3 clusters for the second dataset (Extended Iris dataset). This consistency closely aligns with the established ground truth, indicating a marginal effect of the privacy parameter $\epsilon$ on the server’s cluster count recommendations. However, it is important to note that the actual quality of the clusters formed is subject to the specific clustering algorithm employed. For instance, in the first dataset (Obesity dataset), clustering algorithms demonstrate varied effectiveness influenced by different privacy budgets ($\epsilon$), shown in Table~\ref{tab:effect_of_privacy_budget}. K-Means excel, achieving high ARI values, reaching up to 1.0 when less noise introduced to data (higher $\epsilon$), but maintain low classification accuracy across all settings, indicating well-defined clusters that do not match predefined labels. Silhouette scores also improve with increased $\epsilon$, suggesting clearer cluster definition. Hierarchical Clustering (HC) shows moderate and stable ARI values around 0.48 but face declines in accuracy under extreme privacy settings, hinting at potential misalignments with actual labels. Gaussian Mixture Models (GMM) record lower ARI and negative silhouette scores, suggesting less effective clustering and poor separation, with fluctuating accuracy that sometimes aligned with class labels under minimal privacy constraints. DBSCAN consistently performs poorly with very low ARI, negative silhouette scores, and minimal accuracy, indicating its unsuitability for this dataset due to its sensitivity to specific parameter settings and data density.
In the second dataset (Extended Iris dataset),  the performance of clustering algorithms vary significantly under different privacy settings as shown in Table~\ref{tab:effect_of_privacy_budget_iris}. K-Means showcases excellent clustering with ARI values of 0.997 at low and high $\epsilon$ levels, though it drops at $\epsilon$ = 1, reflecting its sensitivity to privacy settings, despite maintaining high silhouette scores for good cluster separation. However, its consistently low accuracy indicates a misalignment between the clusters and actual class labels. Hierarchical Clustering (HC) remains stable across all metrics and $\epsilon$ settings, achieving moderate to high ARI and silhouette scores, and comparatively better accuracy at 0.38, suggesting it aligns more closely with true labels. Gaussian Mixture Models (GMM) exhibit poor performance with negative ARIs and low silhouette scores, with only moderate accuracy, underscoring its challenges in this dataset under privacy constraints. DBSCAN performs poorly, with extremely low ARI, negative silhouette scores, and zero accuracy across all $\epsilon$ settings, confirming its unsuitability for the dataset. Overall, the K-Means algorithm excels over others when the server's recommendation was \textit{k} = 3, according to various evaluation metrics. 

Furthermore, the server’s recommendations do not significantly deviate from the original data in both datasets. To understand the behavior of data points in dataset \#1, we conducted an analysis by selecting two clusters from the original dataset and applying the RR mechanism with varying $\epsilon$ values. This investigation revealed that the RR mechanism effectively maintains the separation between clusters when present. 

\begin{figure*}[h]
\centering
\begin{subfigure}{0.3\textwidth}
\includegraphics[width=\linewidth]{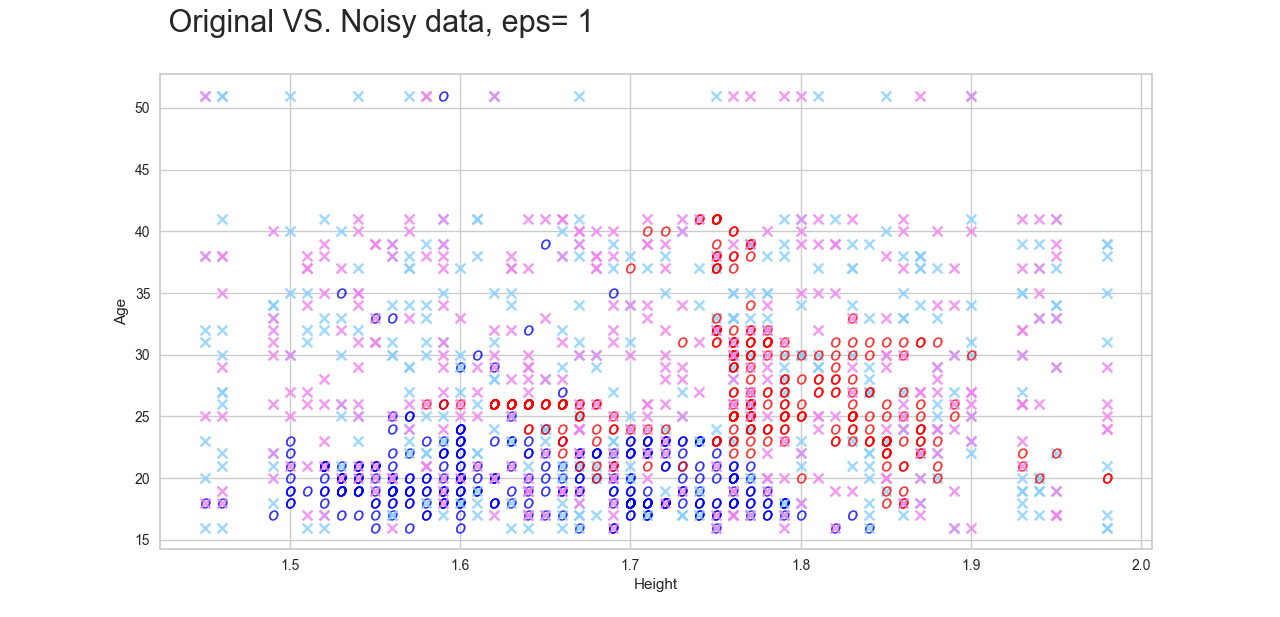}
\includegraphics[width=\linewidth]{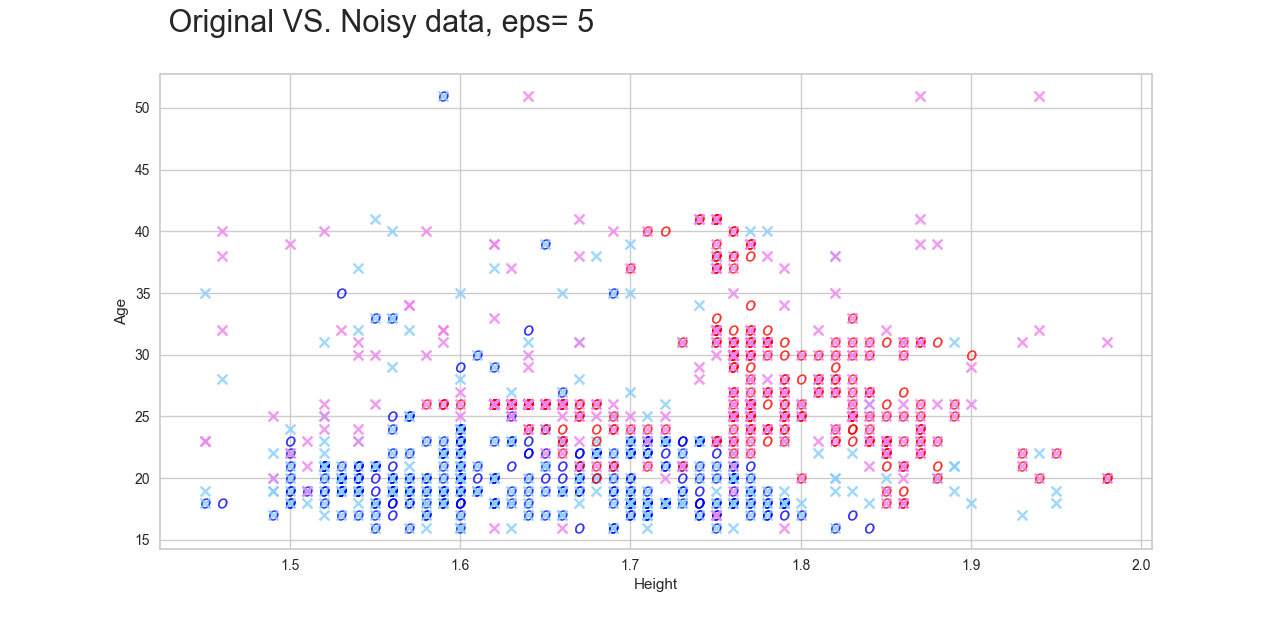}
\includegraphics[width=\linewidth]{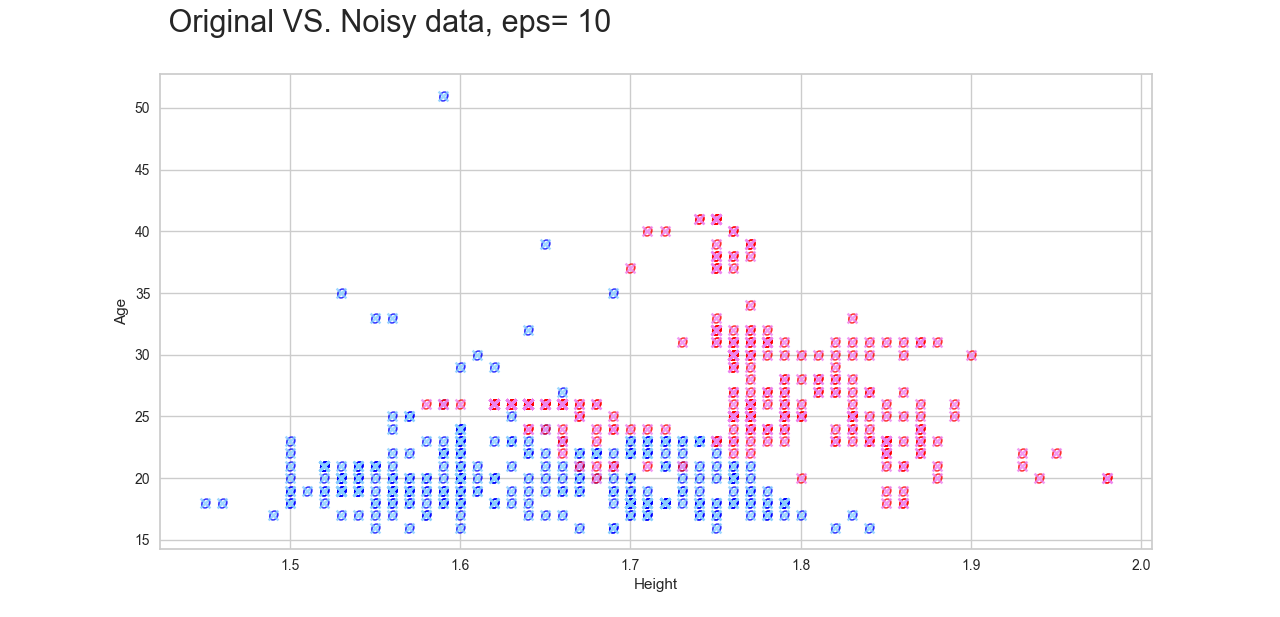}
\caption{}
\label{fig: originalVsNoisy2}
\end{subfigure}
\begin{subfigure}{0.3\textwidth}
\includegraphics[width=\linewidth]{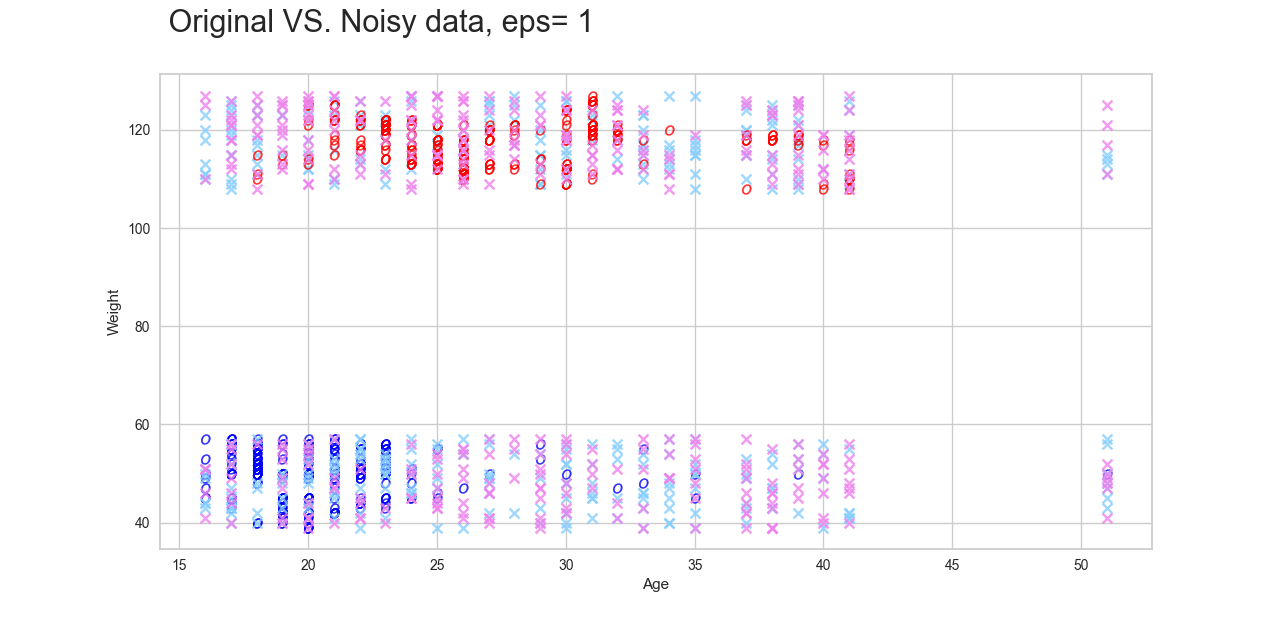}
\includegraphics[width=\linewidth]{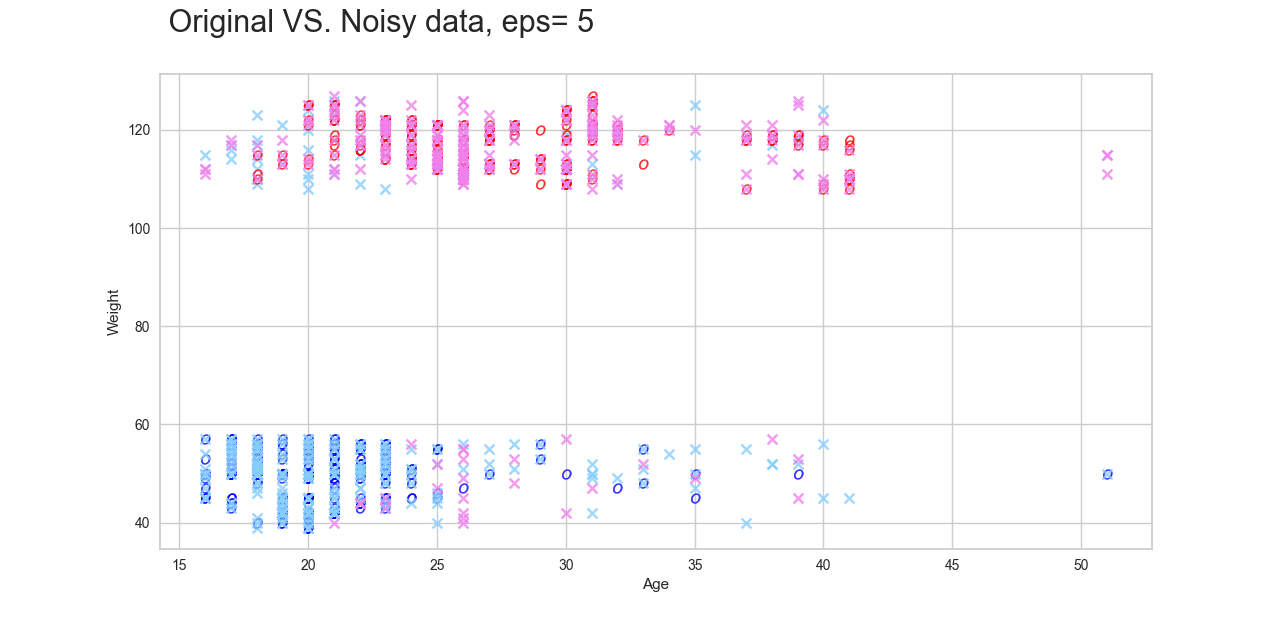}
\includegraphics[width=\linewidth]{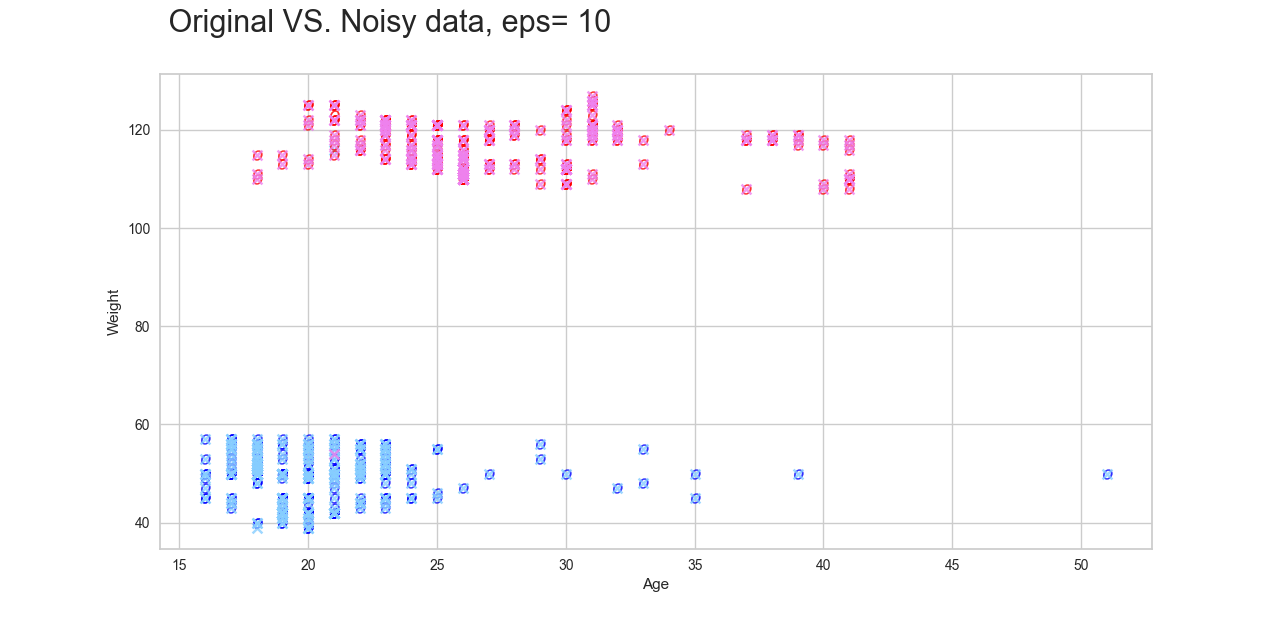}
\caption{}
\label{fig: originalVsNoisy}
\end{subfigure}
\begin{subfigure}{0.3\textwidth}
\includegraphics[height=2.65cm]{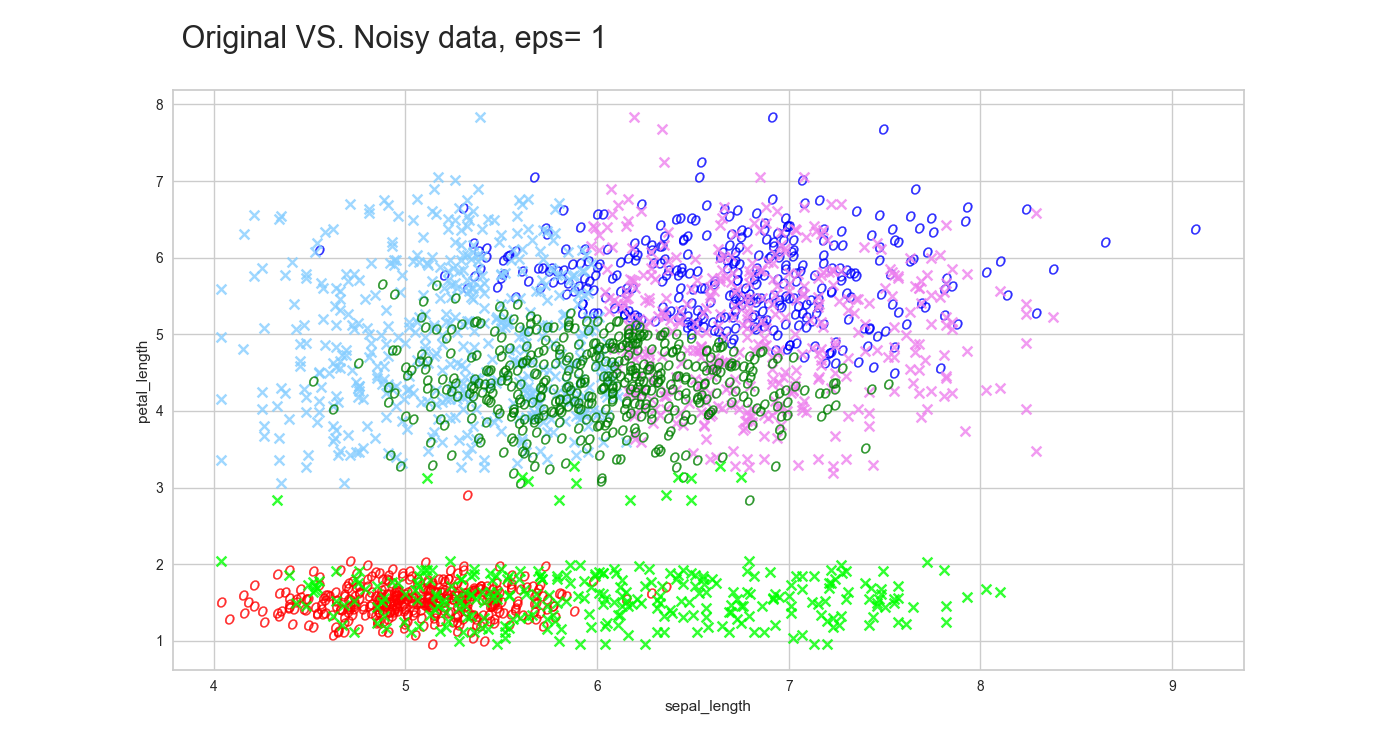}
\includegraphics[height=2.65cm]{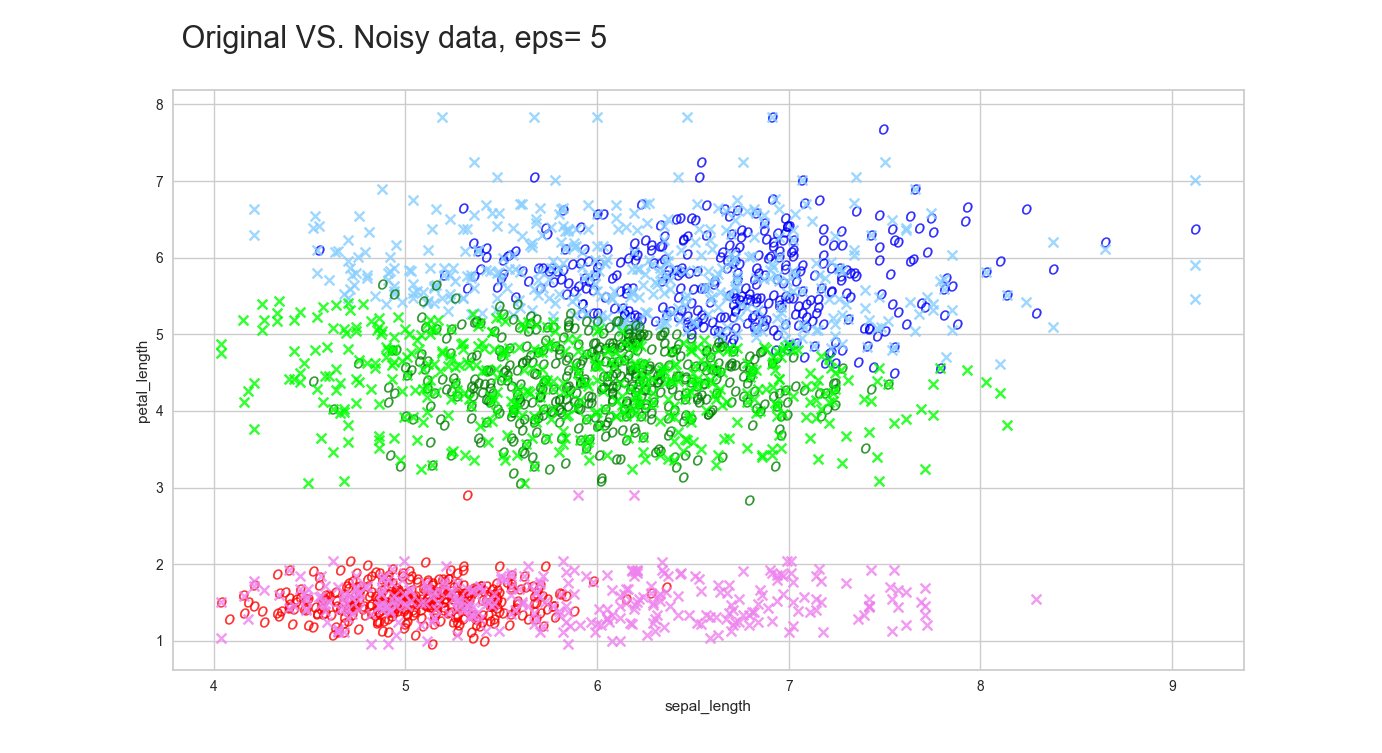}
\includegraphics[height=2.65cm]{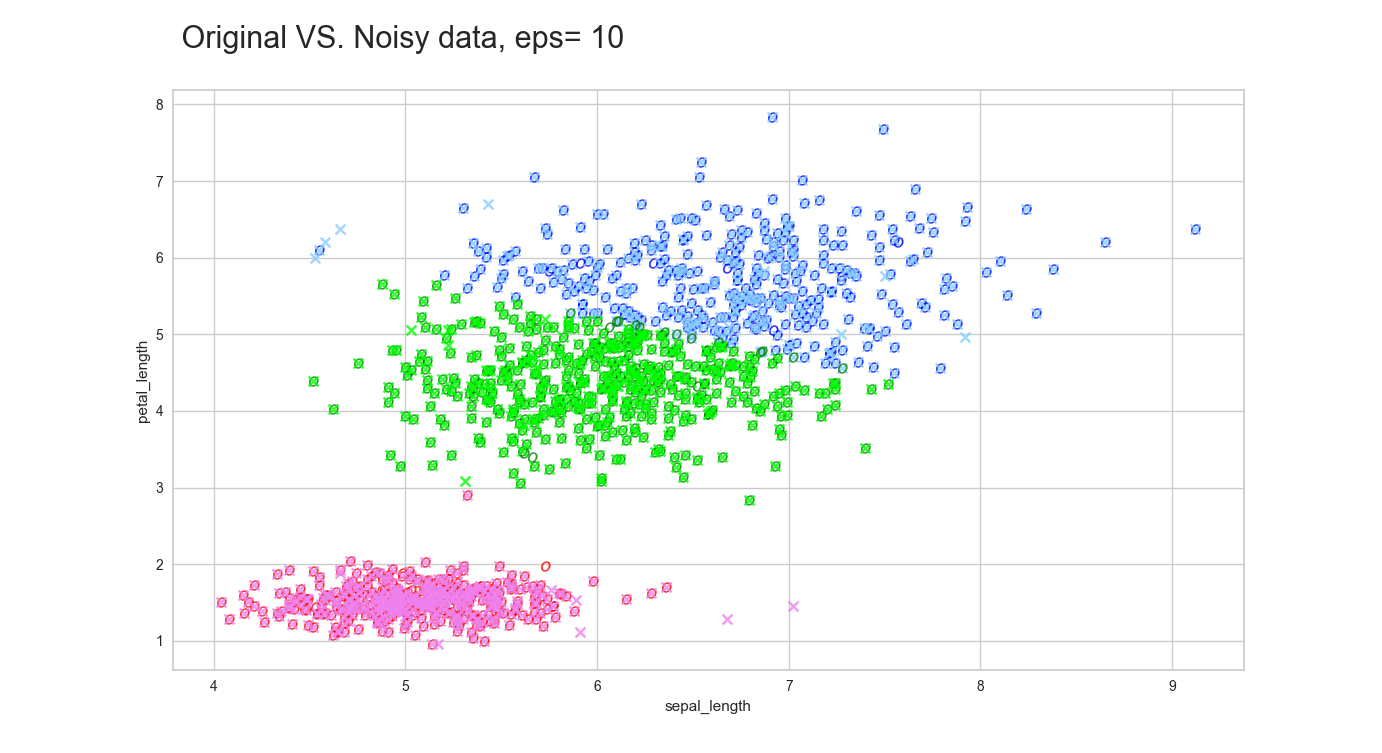}
\caption{}
\label{fig: originalVsNoisy_iris}
\end{subfigure}
\caption{ \textbf{(a)}:Contrast in dataset \#1 with Overlapping Clusters ($\epsilon$ = 1, 5, 10): This part displays the differences between original ('O') and noise-modified ('X') data in closely positioned clusters, colored blue and red. \textbf{(b)} : Comparison in dataset \#1 with Clear Cluster Gaps ($\epsilon$ = 1, 5, 10): Here, the focus is on the impact of the Randomized Response (RR) method on data (original 'O', noisy 'X') in maintaining cluster gaps despite noise variations, balancing privacy with data structure integrity. \textbf{(c)}: Original vs. Noisy Data in dataset \#2 ($\epsilon$ = 1, 5, 10): This section compares original ('O') and noise-affected ('X') data at different privacy levels, using blue, red, and green to show cluster separation effectiveness via the RR mechanism. Note: Plots can be zoomed in for clearer visualization.
}
\label{fig:data_compar}
\end{figure*}

A comparison of Figures~\ref{fig: originalVsNoisy2} and~\ref{fig: originalVsNoisy} illustrates that the distinction between two randomly selected clusters is retained even when the data is subjected to different $\epsilon$ values. This finding is significant as it demonstrates that despite lower $\epsilon$ values possibly leading to a more sparse appearance of the data, the server is still capable of accurately identifying two distinct clusters. This is because the RR mechanism ensures that data points are redistributed within a range akin to their original positions. Furthermore, an analysis of Table~\ref{tab:server_suggestion_10_0.1} shows that the server’s recommendations for second dataset closely mirror the original data. An exploration involving a comparison of the original and RR-perturbed data points across different $\epsilon$ values, as demonstrated in Figure~\ref{fig: originalVsNoisy_iris}, indicates that in two out of the three clusters in dataset \#2, data points overlap without a clear gap, while the third cluster's points are notably distanced from the others. This observation reinforces the notion that the RR mechanism is capable of preserving existing gaps between clusters for various $\epsilon$ values.
These results underscore the RR mechanism's proficiency in safeguarding the intrinsic structure of the data while incorporating elements of privacy protection. By effectively maintaining the relative distances between data points, the server is enabled to provide precise recommendations for the number of clusters, despite the noise caused by different $\epsilon$ values. This highlights the RR mechanism's balance in protecting data privacy while ensuring the accuracy of clustering algorithm suggestions in a privacy-conscious data analysis setting.
\begin{table}[h]
    \caption{Impact of Data Sharing Proportions on Clustering Algorithms' Performance in the dataset \#1. This table evaluates how different proportions of data shared with the server (10\%, 30\%, 50\%) influence the clustering outcomes (ARI, Silhouette, and Accuracy) for various algorithms (K-Means, HC, GMM, DBSCAN) at a fixed privacy parameter ($\epsilon = 0.1$). }
    \centering
    \begin{tabular}{l|c|c|c|c|c|c}
    \hline
         Algorithm & Shared & $\epsilon$ & K  & ARI & Silhouette & Accuracy \\
         \hline
         K-Means & 10\% & 0.1 & k = 8 & 0.75 & 0.41 & 0.18 \\
         K-Means & 30\% & 0.1 & k = 8 & 0.75 & 0.41& 0.18\\
         K-Means & 50\% & 0.1 & k = 8 & 0.75 & 0.41& 0.18\\
         \hline
         HC & 10\% & 0.1 & k = 8 & 0.481 & 0.39 &0.005\\
         HC & 30\% & 0.1 & k = 8 & 0.481 & 0.39 &0.005\\
         HC & 50\% & 0.1 & k = 8 & 0.0.481 & 0.39&0.005 \\
         \hline
         GMM & 10\% & 0.1 & k = 6 & 0.185 & -0.143&0.201 \\
         GMM &30\% & 0.1 & k = 8 & 0.175 & -0.111 &0.18\\
         GMM & 50\% & 0.1 & k = 5 & 0.169 & -0.001&0.23 \\
         \hline
         DBSCAN & 10\% & 0.1 & k = 10 & 0.017 & -0.504& 0.005 \\
         DBSCAN & 30\% & 0.1 & k = 10 & 0.017 & -0.504& 0.005\\
         DBSCAN & 50\% & 0.1 & k = 10 & 0.017 & -0.504 &0.005\\
    \end{tabular}
    \label{tab:effect_of_amount_of_data}
\end{table}

\begin{table}[h]
    \caption{Analysis of Server Recommendations for Clustering Parameters Based on Data Sharing Amounts in the second Dataset. This table examines the influence of varying amounts of data shared (10\%, 30\%, 50\%) on server-suggested clustering parameter (\textit{k}) and their resulting ARI, Silhouette, and Accuracy metrics at a constant privacy parameter ($\epsilon = 0.1$). }
    \centering
    \begin{tabular}{l|c|c|c|c|c|c}
    \hline
         Algorithm & Shared & $\epsilon$ & K  & ARI & Silhouette & Accuracy \\
         \hline
         K-Means & 10\% & 0.1 & k = 3 & 0.997 & 0.52 &0\\
         K-Means & 30\% & 0.1 & k = 2 & 0.44 & 0.57 &0.18\\
         K-Means & 50\% & 0.1 & k = 2 & 0.44 & 0.57 &0.18\\
         \hline
         HC & 10\% & 0.1 & k = 3 & 0.84 & 0.51& 0.38\\
         HC & 30\% & 0.1 & k = 2 & 0.55 & 0.52 &0.66\\
         HC & 50\% & 0.1 & k = 2 & 0.55 & 0.52 &0.66\\
         \hline
         GMM & 10\% & 0.1 & k = 3 & -0.0003 & 0.021 & 0.3\\
         GMM & 30\% & 0.1 & k = 2 & -0.0004 & 0.051 &0.32\\
         GMM & 50\% & 0.1 & k = 2 & -0.0004 & 0.051 &0.32\\
         \hline
         DBSCAN & 10\% & 0.1 &  k = 6 & 0.003 & -0.6 & 0\\
         DBSCAN & 30\% & 0.1 & k = 6 & 0.003 & -0.6 &0\\
         DBSCAN & 50\% & 0.1 &  k = 6 & 0.003 & -0.6 &0\\
    \end{tabular}
    \label{tab:effect_of_amount_of_data_iris}
\end{table}
\subsubsection{Impact of Shared Data Volume on Server Suggestions}
In exploring the influence of shared data volume on clustering algorithm suggestions for both datasets 1 and 2, the results consistently indicate that varying the proportion of data shared with the server does not significantly impact the server's recommendations for clustering input parameters. To investigate this, we conduct experiments where varying amounts of data are shared with the server while keeping the privacy parameter ($\epsilon$) unchanged. This observation is consistent across both datasets and all tested algorithms, as shown in Tables~\ref{tab:effect_of_amount_of_data} and~\ref{tab:effect_of_amount_of_data_iris}.

For the first dataset, the K-Means algorithm maintains the same ARI, Silhouette, and Accuracy metrics across different data sharing proportions, suggesting that its performance remains stable despite changes in the volume of data shared. Similarly, Hierarchical Clustering (HC), Gaussian Mixture Models (GMM), and DBSCAN show consistent performance metrics across different data sharing amounts, further supporting the notion that the quality of clustering recommendations does not deteriorate with reduced data sharing.
In the second dataset, similar patterns emerge. For instance, the K-Means algorithm and HC adjust their suggested number of clusters slightly depending on the data share, but the overall performance metrics such as ARI and Silhouette remain relatively stable. This trend continues with GMM and DBSCAN, which also show little variation in performance across different data sharing proportions.

These findings suggest that the server is capable of providing robust and reliable recommendations for clustering parameters regardless of the amount of data shared, enabling effective clustering outcomes even when data owners choose to share minimal data. This is particularly advantageous in scenarios where data privacy is a concern, as it allows data owners to restrict the amount of shared data without compromising the effectiveness of the clustering process. Overall, the server's ability to consistently suggest appropriate clustering parameters across varying data proportions demonstrates its effectiveness and reliability in guiding the clustering process under different data availability conditions.
\section{Privacy Analysis: Membership Inference Attack}

\begin{figure}[h]
    \centering
    \includegraphics[width = \linewidth]{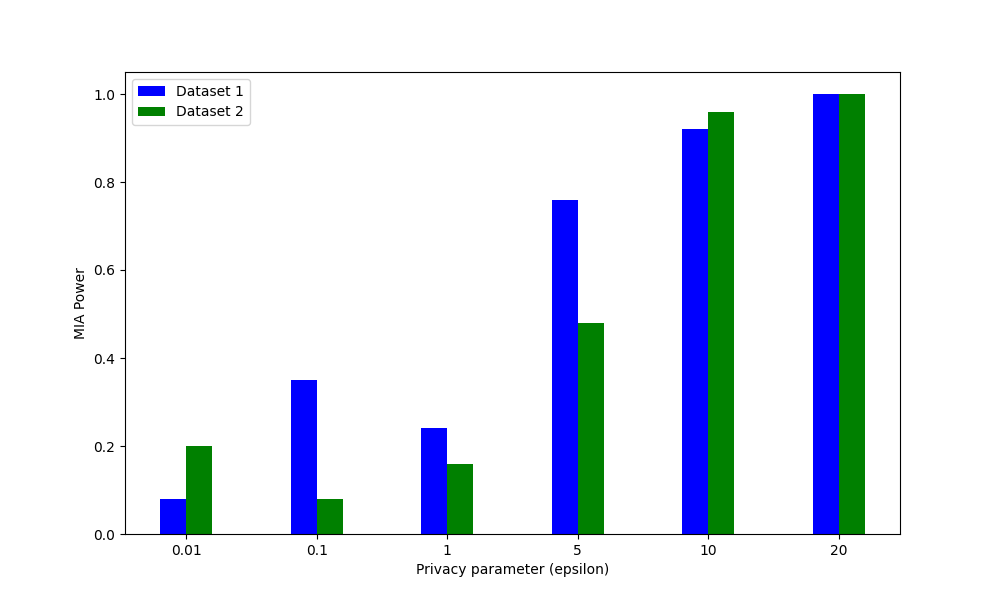}
    \caption{Analysis of Membership Inference Attack Risks: This figure illustrates the increasing likelihood of data identification in two datasets as privacy parameters ($\epsilon$) increase. The blue bars represent dataset \#1, and the green bars represent dataset \#2, highlighting the direct correlation between reduced noise levels and heightened data vulnerability.}
    \label{fig:MIA_results}
\end{figure}

Membership inference attacks (MIA) are techniques used to determine whether specific individual data was included in a dataset. These attacks pose significant privacy risks, especially when datasets contain sensitive information. Our goal is to minimize these risks for individuals whose data is part of a dataset shared with others.
To enhance data privacy, only a portion of the dataset, even in its noisy format, is shared with the server. It has been observed that the likelihood of successful membership inference attacks is inversely related to the amount of noise added to the dataset. We divide the data into two groups to assess the impact of these attacks:

    \textit{Case Group:} This group contains data from specific number of individuals (150 for first dataset and 100 for second dataset) and represents the subset of the dataset that is shared with server, thus exposed to potential membership inference attacks.

    \textit{Control Group:} This group includes data that remains entirely internal and is not shared with the server. It serves as a benchmark to gauge the risk of data exposure.

Our approach to membership inference attacks includes computing a threshold to determine membership based on the degree of similarity between shared and unshared data. This threshold helps to determine whether an individual's data was likely part of the dataset used to train the model.

Following the threshold determination, we assess each individual's data against this threshold to ascertain their membership. We calculate the similarity between each individual's data in the case group and the control group using distance measures. If the similarity exceeds the threshold, the data point is considered at risk of exposure through membership inference.    

Our detailed analysis is visually represented in Figure~\ref{fig:MIA_results}, demonstrating the impact of the privacy parameter ($\epsilon$), with increased $\epsilon$ values reducing the noise and thereby increasing the risk of data identification.

For both datasets, the findings show that as $\epsilon$ increases, the risk of successful membership inference attacks also rises. This suggests that less noise in the data leads to higher likelihoods of identifying individual data points within the dataset.

These findings underscore the critical need to account for privacy concerns regarding membership inference attacks when sharing data. Limiting the amount of shared data and augmenting it with noise is a vital strategy to reduce the likelihood of such attacks, thereby preserving the privacy of individuals within the dataset. This necessitates a strategic approach in the protocols of data sharing and the implementation of privacy-enhancing techniques, which are crucial in collaborative clustering processes. It highlights the necessity of balancing data utility with privacy considerations to safeguard sensitive information effectively.

\section{Conclusion}
The primary goal of this study is to identify the most suitable input parameters for four widely-used clustering algorithms. The study aims to facilitate collaborative clustering tasks between multiple data owners, simplifying its experiments within a two-party context. To enhance the reliability and accuracy of the clustering process, the framework introduces a semi-trusted third party. This third party plays a crucial role in offering informed recommendations concerning the optimal clustering algorithm and the associated input parameters, with a particular emphasis on maximizing utility and performance. The results demonstrate that neither the quantity of perturbed data shared with the third party (referred to as the server), nor the privacy budget ($\epsilon$), has a significant impact on the server's suggestions.

Furthermore, this study conducts an analysis of membership inference attacks to evaluate the vulnerability of the system. As the privacy budget ($\epsilon$) increases, the power of membership inference attacks also increases. This indicates that higher levels of privacy budget compromise the effectiveness of privacy protection, making it easier for attackers to infer whether an individual's data is part of the shared dataset.

These findings emphasize the need for careful consideration of privacy-preserving mechanisms and the importance of maintaining an appropriate balance between privacy protection and utility. While the server's suggestions for input parameters remain consistent regardless of the amount of perturbed data or the privacy budget, the potential risks associated with membership inference attacks highlight the need to adopt appropriate safeguards and mitigation strategies. Protecting the privacy of individuals and ensuring the security of collaborative clustering processes should be key priorities in future research and system design.

%
%

\bibliographystyle{IEEEtran}
\end{document}